\pdfoutput=1

\documentclass[11pt]{article}

\usepackage{authblk}
\usepackage[final]{acl}
\usepackage{relsize}
\usepackage{times}
\usepackage{latexsym}
\usepackage{tabularx}
\usepackage{enumitem}
\usepackage{algorithm}
\usepackage{booktabs}
\usepackage{multirow}
\usepackage{adjustbox}
\usepackage{algpseudocode}
\usepackage{listings}
\usepackage{xcolor}

\usepackage{booktabs} 
\usepackage[T1]{fontenc}

\usepackage[utf8]{inputenc}

\usepackage{microtype}

\usepackage{inconsolata}
\usepackage{amsmath}
\usepackage{amsfonts}
\usepackage{footnote}
\usepackage{booktabs}
\usepackage{hyperref} 
\usepackage{multirow}
\usepackage{colortbl}
\usepackage{xcolor}
\usepackage{subfigure}

\usepackage{graphicx}
\usepackage{makecell}

\begin{document}

%
%

\title{Emergent Convergence in Multi-Agent LLM Annotation}



\author[1,2]{\textbf{Angelina Parfenova}}
\author[1]{\textbf{Alexander Denzler}}
\author[2]{\\\textbf{Juergen Pfeffer}}
\affil[1]{Lucerne University of Applied Sciences and Arts}
\affil[2]{Technical University of Munich}
\maketitle
\renewcommand{\thefootnote}{\fnsymbol{footnote}}

\begin{abstract}
Large language models (LLMs) are increasingly deployed in collaborative settings, yet little is known about how they coordinate when treated as black-box agents. We simulate 7,500 multi-agent, multi-round discussions in an inductive coding task, generating over 125,000 utterances that capture both final annotations and their interactional histories. We introduce process-level metrics—code stability, semantic self-consistency, and lexical confidence—alongside sentiment and convergence measures, to track coordination dynamics. To probe deeper alignment signals, we analyze the evolving geometry of output embeddings, showing that intrinsic dimensionality declines over rounds, suggesting semantic compression. The results reveal that LLM groups converge lexically and semantically, develop asymmetric influence patterns, and exhibit negotiation-like behaviors despite the absence of explicit role prompting. This work demonstrates how black-box interaction analysis can surface emergent coordination strategies, offering a scalable complement to internal probe-based interpretability methods.
\end{abstract}

\section{Introduction}

Inductive coding is a core method in qualitative research, used to identify patterns and themes by assigning semantic labels, or \textit{codes}, to unstructured text segments \cite{Saldana2016, Braun2021}. This process is typically carried out by human coders who iteratively interpret, categorize, and refine codes. Collaborative coding can enhance interpretive depth through discussion and consensus, but it is also time-consuming and subject to inconsistencies caused by individual bias and group effects \cite{MacQueen2008, Bernard2016, bumbuc2016subjectivity}.

Recent advances in large language models (LLMs) have created opportunities to automate parts of the qualitative analysis pipeline. While prior work has examined LLMs for individual coding \cite{chen2024computational, parfenova-etal-2025-text}, little is known about how they behave in \textit{multi-agent} settings that mirror human annotation teams. In particular, it remains unclear how coordination arises between models, and whether convergence in their outputs reflects shared semantic understanding, lexical mimicry, or other surface-level alignment processes.

\begin{figure}
    \centering
    \includegraphics[width=\linewidth]{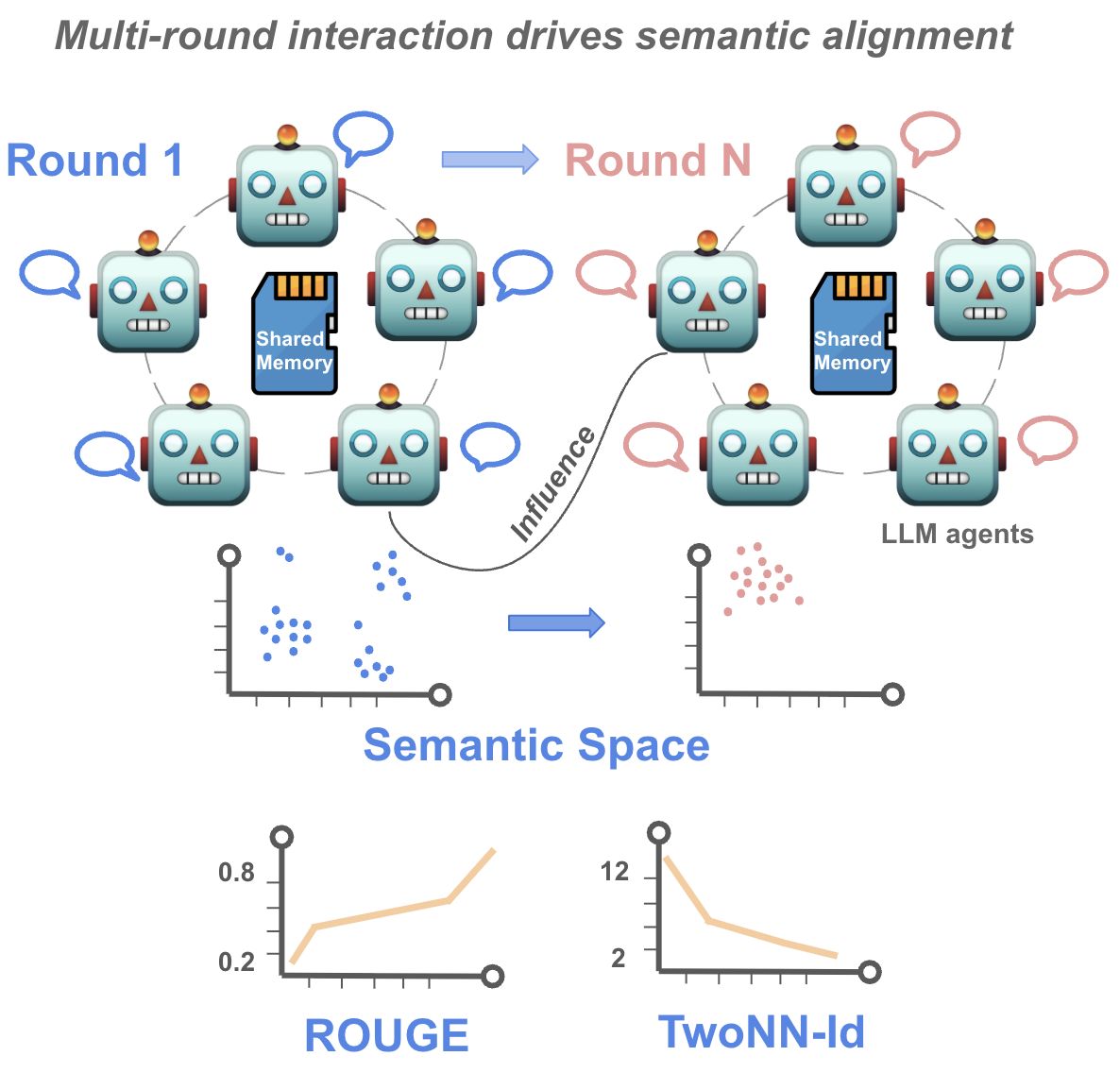}
    \caption{Overview of our multi‑agent simulation framework. LLM agents iteratively exchange outputs via a shared conversational memory, progressing from Round~1 to Round~$N$. Over rounds, codes move from dispersed to clustered in semantic space, while ROUGE increases and intrinsic dimensionality (TwoNN‑Id) decreases, indicating lexical convergence and semantic compression.}

    \label{fig:baner}
\end{figure}

This paper introduces a large-scale simulation framework for multi-agent, multi-round LLM discussions in an inductive coding task. Each model acts as a black-box agent that proposes and revises codes over several rounds, without finetuning or access to internal states. By analyzing only the outputs, we track how codes evolve and align over time, providing an interpretable view of coordination.

We combine process-level metrics with geometric analysis of output embeddings to capture both lexical and structural aspects of coordination. Our metrics include \textit{code stability}, \textit{semantic self-consistency}, and \textit{opinion–confidence} dynamics, alongside measures of embedding-space geometry such as intrinsic dimensionality, which we interpret as a proxy for semantic compression during convergence.

Our contributions are:
\begin{enumerate}
    \item We present a simulation framework for large-scale, black-box multi-agent LLM discussions applied to collaborative annotation.
    \item We propose coordination metrics that capture surface-level stability, semantic consistency, and opinion–confidence alignment, as well as geometric shifts in embedding space.
    \item We empirically demonstrate that multiround interactions enhance lexical convergence, reduce embedding space dimensionality, and yield asymmetric influence patterns between models.
\end{enumerate}



\section{Background}

Qualitative data analysis (QDA) is a central method in the social sciences, used to identify and interpret patterns in unstructured text \cite{miller1990introduction, Creswell2016}. A core step in QDA is coding, where analysts assign short labels to data segments to capture their essential meaning \cite{Saldana2016}. These \textit{codes} form the building blocks for higher-level categories and themes \cite{Braun2021}. Coding is often conducted by teams of human analysts who iteratively refine code definitions and resolve disagreements through discussion.

Recent work has explored the use of large language models (LLMs) to assist or automate parts of the coding process \cite{Bommasani2021, Fabbri2021, parfenova-etal-2024-automating}. While LLMs can offer gains in speed and consistency, concerns remain about the reliability and interpretability of their outputs, particularly for tasks requiring subjective judgment \cite{Morse2017, MacQueen2008, bumbuc2016subjectivity, Bernard2016}.

Human coding is not only a technical activity but also a \textit{social one}, shaped by negotiation, influence, and consensus-building. Research in social psychology shows that group decisions are affected by factors such as confidence, majority opinion, and perceived commitment \cite{moussaid2013social, suzuki2015neural}. These findings suggest that group coding dynamics extend beyond individual annotation to include broader mechanisms of coordination and persuasion.

In parallel, studies have modeled LLM agents in multi-party negotiation and collaboration settings, showing that they can replicate aspects of human interaction, including persuasion strategies and iterative reasoning \cite{fu2023improvinglanguagemodelnegotiation, deng2024llms, abdelnabi2024cooperationcompetitionmaliciousnessllmstakeholders}. For example, \citet{vaccaro2025advancingainegotiationsnew} show that negotiation outcomes among LLM agents are influenced by social dimensions such as warmth and dominance, and that LLMs exhibit reasoning strategies not fully predicted by existing human negotiation theories.

However, these interactional capabilities have not been systematically examined in applied annotation settings such as qualitative coding, where sustained collaborative interpretation is central. Existing research has largely focused on single-turn or task-specific interactions, leaving open questions about how LLMs behave in multi-turn, group-based annotation, and how such interactions shape the semantic and structural properties of their outputs.

This work addresses this gap by simulating multi-agent, multi-round discussions among diverse LLMs performing inductive coding. Drawing on insights from qualitative research, social psychology, and agent-based modeling, we analyze how LLMs negotiate semantic content, influence each other’s decisions, and converge—or fail to converge—on shared annotations.

\section{Dataset}

We construct a dataset for collaborative qualitative coding by sampling 500 English-language comments from the Jigsaw Unintended Bias in Toxicity Classification dataset.\footnote{\url{https://www.kaggle.com/c/jigsaw-unintended-bias-in-toxicity-classification}} Comments are selected based on two criteria: (1) high annotator disagreement scores to capture subjectivity, and (2) a minimum length of 100 words to ensure interpretive richness. The resulting dataset consists of identity-related, linguistically diverse utterances suited for thematic analysis.

\section{Experimental Setup}
\label{sec:experimental_setup}
We simulate structured group discussions among large language models (LLMs), varying both group size (2, 3, or 5 agents) and discussion depth (1–5 rounds). For each configuration, we generate 500 discussions, yielding a total of 7,500 multi-agent simulations. Each discussion proceeds through three phases:
(1) initial code generation from each agent,
(2) one to five rounds of turn-based refinement, and
(3) final synthesis.

Agents take turns in a fixed sequence. After each turn, the agent is prompted to summarize its message in a single sentence. These summaries are accumulated as conversational memory and provided as context for subsequent responses, approximating memory through accumulated turn summaries. This setup avoids external memory modules and finetuning, instead relying on prompt-based inference alone. It also enables scalable context management while preserving discourse coherence. The full algorithm is outlined in Algorithm \ref{alg:multiturn} (Appendix A). 

Each discussion is initialized with one of five prompt templates (see Table~\ref{tab:prompts}), ranging from formal coding instructions to informal summary requests. The prompts are iterated sequentially to ensure balanced coverage.

\begin{table}[h]
\centering
\tiny
\begin{tabular}{ll}
\toprule
\textbf{Prompt ID} & \textbf{Prompt Text} \\
\midrule
1 & A code is often a word or short phrase that symbolically assigns a \\
  & salient, essence-capturing and/or evocative attribute to a portion \\
  & of language-based or visual data. Perform thematic analysis on the \\
  & following comment and generate a concise qualitative code. \\
2 & Summarize the main idea of this sentence in a short, thematic code. \\
3 & From the perspective of a social scientist, summarize the following \\
  & sentence as you would in thematic coding. \\
4 & Can you tell me what the main idea of this sentence is in just a few words? \\
5 & If you were a social scientist doing thematic analysis, what code \\
  & would you give to this citation? \\
\bottomrule
\end{tabular}
\caption{Prompt formulations used across the simulation setup to elicit qualitative codes from models \cite{parfenova-etal-2025-text}. Each discussion is seeded with one of these prompts.}
\label{tab:prompts}
\end{table}

\subsection{Corpus Statistics}

Across all settings, the simulation produces 7,500 discussions. With each agent generating an initial code, multiple turn-level refinements, and a final synthesis, the corpus contains approximately 125,000 discrete agent utterances. We store all transcripts, turn summaries, and final codes in structured JSON and CSV formats for reproducible analysis.

\section{Metrics}

We compute ROUGE-1, ROUGE-2, and ROUGE-L scores \cite{lin-2004-rouge} to quantify lexical similarity and convergence between generated codes over discussion rounds. To inspect structural patterns in embedding space, we project sentence embeddings with UMAP \cite{mcinnes2018umap}, visualizing both 2D and 3D clustering to assess inter-model separability and convergence.

We also evaluate toxicity using the Unitary Toxicity classifier,\footnote{\url{https://huggingface.co/unitary/unbiased-toxic-roberta}}, treating full agent turns as the unit of analysis.

To capture richer stylistic and psycholinguistic patterns, we apply the ELFEN toolkit\footnote{\url{https://github.com/mmmaurer/elfen}}, extracting features such as lexical diversity, syntactic complexity, and emotional intensity.

\subsection{Stability, Consistency, and Confidence}
We introduce three process-level metrics:

\textbf{Code Stability} – proportion of string-identical outputs between consecutive rounds, reflecting how much each model revises its own prior outputs.

\textbf{Self-consistency Score} – average cosine similarity between TF-IDF representations of a model’s outputs at round $t$ and $t{+}1$, measuring semantic drift.

\textbf{Confidence}
We use a lightweight, output-only proxy for expressed confidence: the length-normalized difference between counts of \emph{certainty} cues (e.g., \textit{clearly, definitely, must}) and \emph{hedging} cues (e.g., \textit{might, possibly, seems}). Formally,
\[
\text{Conf}(u) \;=\; \frac{|C(u)| - |H(u)|}{\text{tokens}(u)}\,.
\]
This does \emph{not} measure true certainty; it approximates how assertive the \emph{language} reads. In Limitations, we discuss sensitivity to lexicon design and normalization (tokens vs.\ words vs.\ sentences).

\begin{figure}
    \centering
    \includegraphics[width=\linewidth]{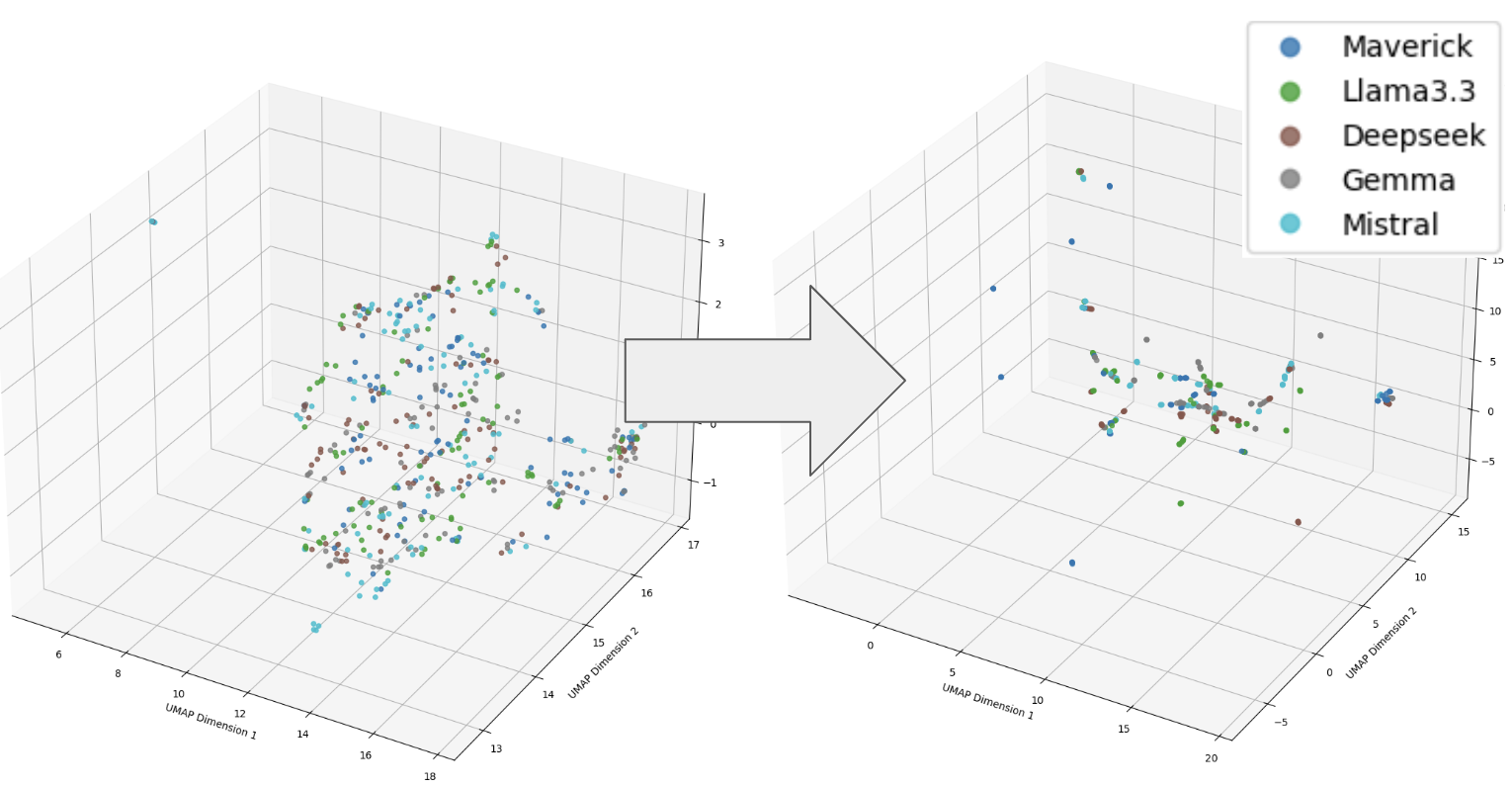}
    \caption{UMAP projection of LLM-generated codes before and after four rounds of multi-agent discussion (5 models). Each point represents a single code, colored by model type. Pre-discussion codes are more dispersed in embedding space (left), while post-discussion codes form tighter clusters with greater cross-model overlap (right). This reflects both lexical convergence and a form of semantic compression, where diverse initial proposals collapse into a lower-dimensional, more aligned representation.}
    \label{fig:umap3d}
\end{figure}

\section{Code Convergence and Quality}

We organize our analysis around two dimensions: the \textit{codes} produced by LLMs at each round, and the \textit{discussion dynamics} that shape those codes. Results span all group sizes (2, 3, or 5 models), rounds (1--5), and prompt types (5 total), highlighting trends in convergence, confidence, and coordination.

\begin{figure}
    \centering
    \includegraphics[width=\linewidth]{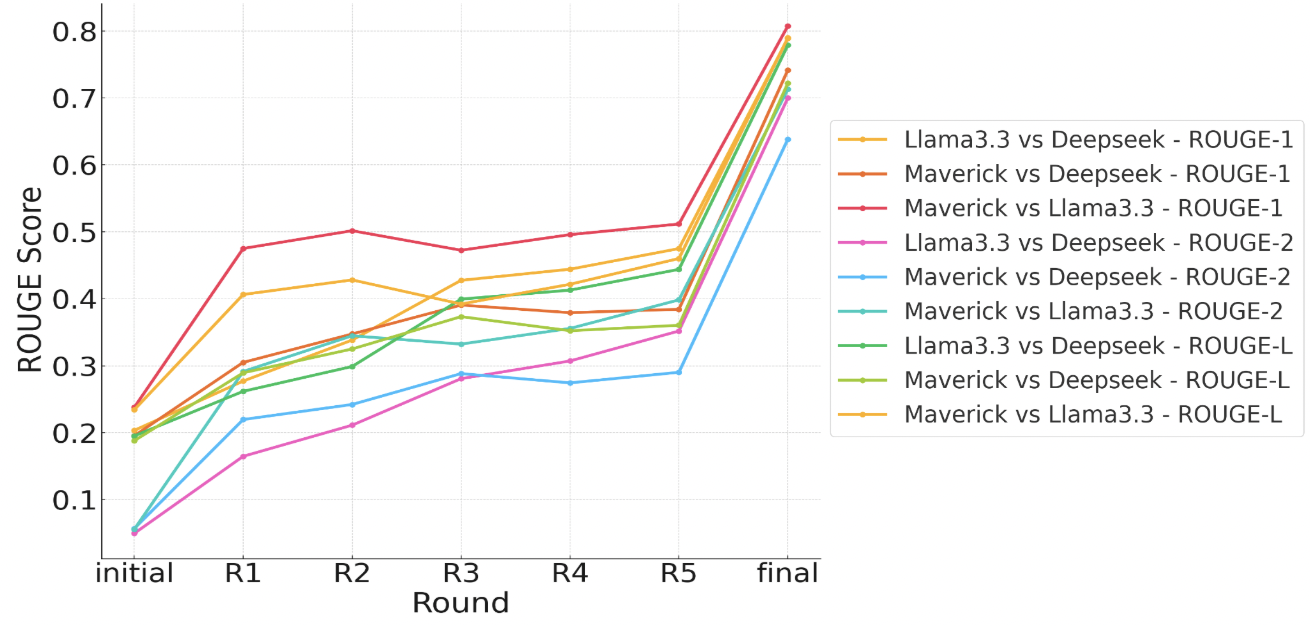}
    \vspace{2pt}
    \caption{ROUGE Score Convergence Across Rounds for Three LLMs. This plot shows ROUGE-1, ROUGE-2, and ROUGE-L similarity scores between pairs of models (Llama4 Maverick, Llama3.3 70B, and Deepseek-R1 70B) across successive discussion rounds. Scores are computed based on model-generated codes at each round, capturing convergence in lexical overlap over time. }
    \label{fig:rougescores}
\end{figure}

\paragraph{ROUGE Analysis.}  
In all configurations, the ROUGE-1, ROUGE-2, and ROUGE-L scores between LLM steadily increase over successive rounds, indicating progressive lexical convergence during the discussion (Figure~\ref{fig:rougescores}). The largest gains occur between the penultimate and final rounds, suggesting that multi-turn interaction yields cumulative benefits that are most visible late in the process. Table~\ref{tab:rouge_l_summary} reports final round ROUGE-L scores across prompts. Peak performance is observed in the 3-model, 4-round setting for \textit{Prompt~1} (Max: 0.8070), with similarly strong outcomes for \textit{Prompt~2} and \textit{Prompt~4}. While additional rounds generally enhance convergence, improvements plateau after the fourth round.

\paragraph{Opinion and Confidence.} Figure~\ref{fig:confidence} (see Appendix B) tracks the average confidence of each model over rounds, estimated from lexical certainty and hedging cues. Mistral produces the most assertive outputs, while Deepseek is the most hedging-prone. All models increase in confidence over time, consistent with growing certainty as the discussion unfolds. 

In Deepseek’s case, the lower confidence scores may be partly attributable to the presence of its reasoning part after \texttt{<think>} token in outputs, which was deliberately not removed during preprocessing and may be interpreted as a hedging or reflective cue by our metric.

\vspace{2pt}

\begin{figure*}
    \centering
    \includegraphics[width=\linewidth]{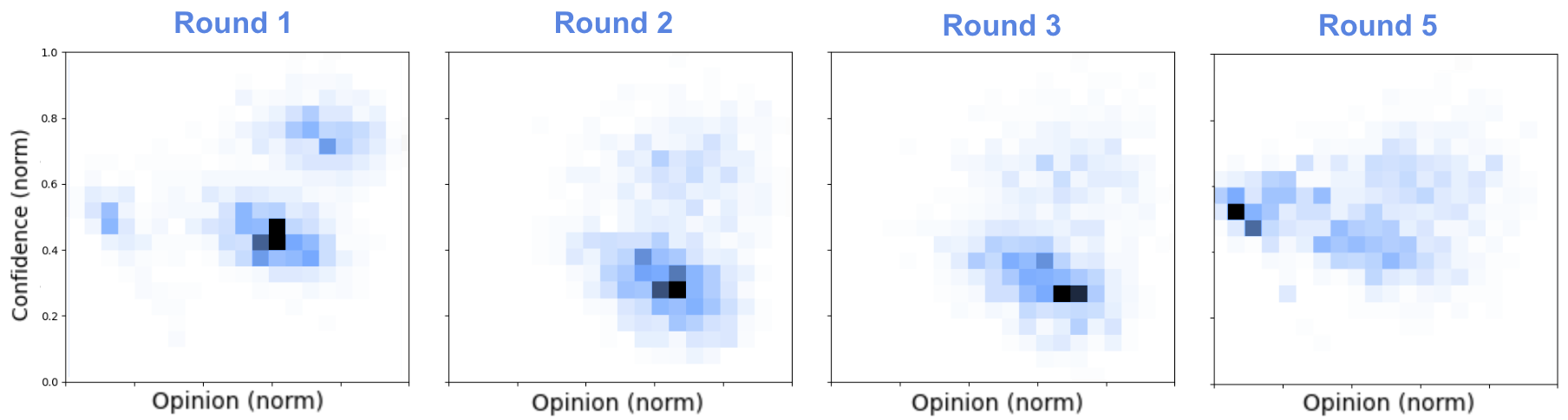}
    \caption{Density plots of normalized opinion vs. confidence across discussion rounds.
Each subplot represents a 2D histogram of model utterances in a given round, showing how expressed opinions (x-axis) relate to confidence scores (y-axis). Darker regions indicate higher concentration of utterances. Over rounds, the distribution evolves from distinct opinion-confidence clusters (Round 1) to more dispersed and overlapping patterns (Rounds 3–5).}
    \label{fig:heatmaps}
\end{figure*}

We further project utterance embeddings into a 2D \textit{opinion–confidence} space. Opinion is derived by reducing 384-dimensional MiniLM sentence embeddings to a single principal component, capturing the primary axis of semantic variance across all responses. Confidence is computed as the normalized frequency difference between certainty and hedging expressions. Heatmaps in Figure~\ref{fig:heatmaps} show that early rounds produce multiple dense clusters (divergent stances, moderate confidence), whereas later rounds reveal consolidation into more concentrated regions, reflecting both semantic and epistemic alignment.

This visualization is inspired by prior social science work on human group decision-making, notably \cite{moussaid2013social}, which mapped individuals into an opinion–confidence space to study convergence under social influence. Our results mirror their findings: multi-round interactions drive both opinion convergence and an overall upward shift in expressed confidence.

\paragraph{Toxicity.} 
Average toxicity scores, measured via a zero-shot social bias detector, generally decrease over time (see Figure~\ref{fig:toxic}, Appendix B). Mistral and Gemma converge to near-zero toxicity by Round~4, while Deepseek maintains relatively higher levels.

\begin{figure}
    \centering
    \includegraphics[width=\linewidth]{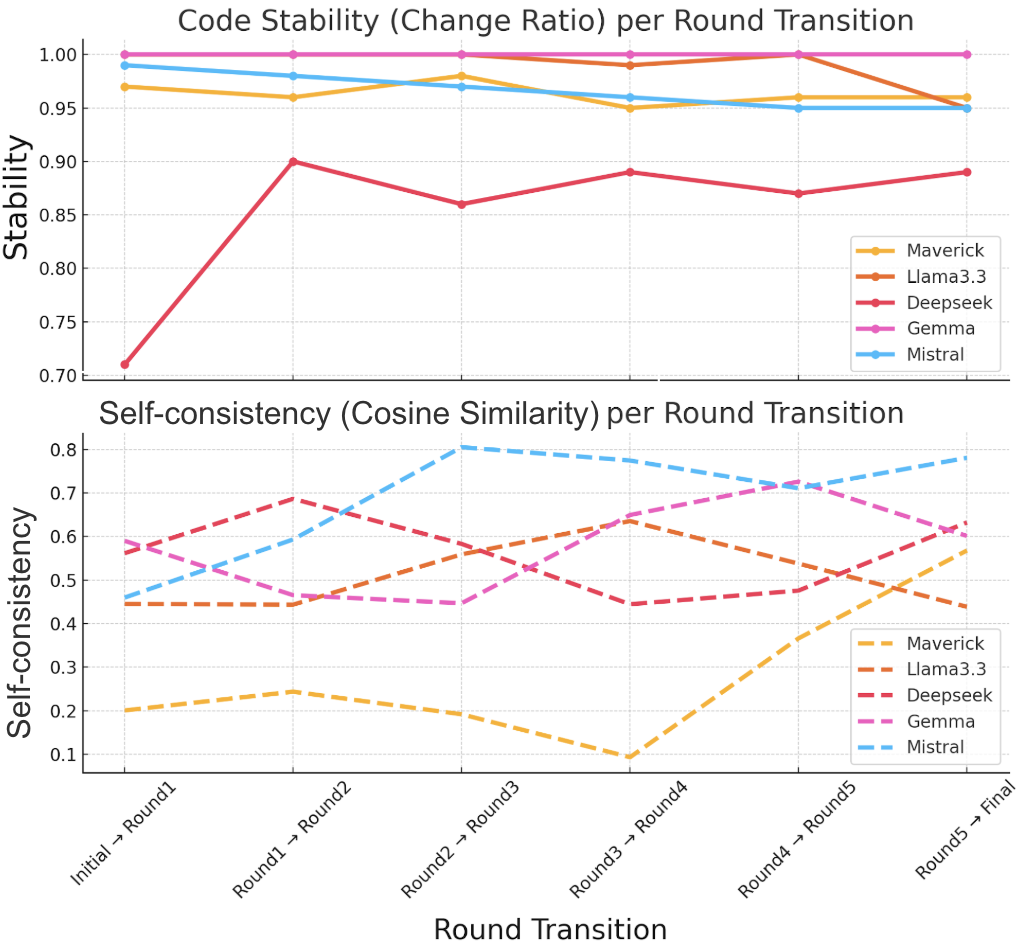}
    \caption{Model Stability and Self-consistency Across Rounds.
Top: Code Stability (1 - change rate) measures the proportion of tokens retained between rounds, reflecting how much models revise their outputs. Bottom: Self-consistency Score is computed as the cosine similarity of TF-IDF vectors between consecutive rounds, indicating semantic consistency. }
    \label{fig:influence}
\end{figure}

\paragraph{Stability and Consistency.} Figure~\ref{fig:influence} summarizes two intra-model metrics: \textit{stability} (percentage of unchanged tokens between rounds) and \textit{self-consistency} (semantic similarity of consecutive outputs). Four models maintain high stability throughout, while Deepseek shows greater variability. Deepseek and Mistral achieve the highest self-consistency, whereas Maverick exhibits more exploratory behavior before converging in later rounds.

\begin{table*}[t]
\centering
\tiny
\caption{Average and Maximum ROUGE-L scores at final round across prompts, number of models, and number of rounds.}
\label{tab:rouge_l_summary}
\begin{adjustbox}{width=\textwidth}
\begin{tabular}{llccccc}
\toprule
\textbf{Prompt} & \textbf{\# Models / Rounds} & \textbf{1} & \textbf{2} & \textbf{3} & \textbf{4} & \textbf{5} \\
\midrule

\multirow{3}{*}{Prompt 1} 
  & 2 models & Avg: 0.7658 / Max: 0.7658 & 0.5124 / 0.5124 & 0.6327 / 0.6327 & 0.7754 / 0.7754 & 0.6851 / 0.6851 \\
  & 3 models & 0.6881 / 0.7089 & 0.7274 / 0.7410 & 0.7409 / 0.8055 & \textbf{0.7729 / 0.8070} & 0.7634 / 0.7897 \\
  & 5 models & 0.6175 / 0.6827 & 0.6450 / 0.7220 & 0.6861 / 0.7410 & 0.7237 / 0.7857 & 0.7062 / 0.7824 \\

\midrule
\multirow{3}{*}{Prompt 2} 
  & 2 models & 0.4703 / 0.4703 & 0.4755 / 0.4755 & 0.5450 / 0.5450 & 0.6274 / 0.6274 & 0.5342 / 0.5342 \\
  & 3 models & 0.5971 / 0.6260 & \textbf{0.7043 / 0.7584} & 0.6728 / 0.7341 & 0.6827 / 0.7463 & 0.6871 / 0.7489 \\
  & 5 models & 0.5225 / 0.6253 & 0.5379 / 0.6370 & 0.5343 / 0.6284 & 0.5677 / 0.6928 & 0.5986 / 0.6961 \\

\midrule
\multirow{3}{*}{Prompt 3} 
  & 2 models & 0.3609 / 0.3609 & 0.3407 / 0.3407 & 0.4056 / 0.4056 & \textbf{0.4193} / 0.4193 & 0.4081 / 0.4081 \\
  & 3 models & 0.2836 / 0.3200 & 0.3417 / 0.4070 & 0.3812 / 0.4941 & 0.3643 / 0.4521 & 0.3950 / 0.4767 \\
  & 5 models & 0.2819 / 0.3866 & 0.3055 / 0.4287 & 0.3306 / \textbf{0.4764} & 0.3194 / 0.4467 & 0.3332 / 0.4658 \\

\midrule
\multirow{3}{*}{Prompt 4} 
  & 2 models & 0.4811 / 0.4811 & 0.4680 / 0.4680 & 0.5185 / 0.5185 & 0.5688 / 0.5688 & 0.4836 / 0.4836 \\
  & 3 models & 0.6880 / 0.7304 & 0.6534 / 0.6768 & 0.7068 / 0.7518 & \textbf{0.7152 / 0.7575} & 0.6363 / 0.6682 \\
  & 5 models & 0.5207 / 0.6544 & 0.5822 / 0.6850 & 0.5932 / 0.6894 & 0.6171 / 0.6688 & 0.5966 / 0.7061 \\

\midrule
\multirow{3}{*}{Prompt 5} 
  & 2 models & 0.5477 / 0.5477 & 0.4712 / 0.4712 & 0.5979 / 0.5979 & \textbf{0.6586 / 0.6586} & 0.5908 / 0.5908 \\
  & 3 models & 0.4417 / 0.4983 & 0.4665 / 0.5789 & 0.4809 / 0.5910 & 0.5102 / 0.5872 & 0.5350 / 0.6474 \\
  & 5 models & 0.3457 / 0.5164 & 0.3602 / 0.5305 & 0.3508 / 0.4881 & 0.3758 / 0.5280 & 0.3975 / 0.5857 \\

\bottomrule
\end{tabular}
\end{adjustbox}
\end{table*}

\paragraph{Semantic Influence Between Models.}
We compute round-wise pairwise cosine similarities between each model’s output and every other model’s prior-round output, producing a 5×5 influence matrix per round (Figure~\ref{fig:semantic-influence}, see Appendix C). Diagonal entries capture self-consistency and off-diagonals capture cross-model semantic alignment.

Early rounds show diffuse influence, with Gemma and Deepseek serving as semantic anchors. By mid-discussion, Llama3.3 emerges as a stronger source of influence, particularly for Gemma and Mistral. Deepseek increasingly absorbs content from peers, acting as a semantic integrator. By Rounds~6–7, both self- and cross-influence scores rise across all models, reflecting a shift toward mutual alignment and emergent inter-model coordination.

\section{Geometric Interpretability of Code Evolution}
\label{sec:geometric}

Recent work by \citet{lee2025geometricsignaturescompositionalitylanguage} has shown that \textit{intrinsic dimensionality} (Id) of neural activations can reveal how language models compress or preserve information during reasoning and instruction following. While their approach uses hidden states, our focus is strictly \textit{output-level geometry}: we track how the intrinsic dimension of LLM-generated codes changes over multi-round discussions, using external sentence embeddings from \texttt{sentence-transformers}. This makes our analysis a \textit{proxy} for semantic complexity in outputs rather than a direct probe of internal representations.

\subsection{Intrinsic Dimensionality Shrinkage Across Rounds}

We estimate Id using the Two-Nearest Neighbor method (TwoNN-Id; \citealt{facco2017estimating}) for codes at each round in 2-, 3-, and 5-model setups (not diving by prompts used). Intuitively, for each point we look at the distances to its nearest and second-nearest neighbors; the distribution of the ratio of these distances has a closed-form dependence on the underlying manifold dimension. Fitting this distribution yields an ID estimate. In our \emph{output-only} setting, ID serves as a proxy for how dispersed or compressed the semantic space of codes is; lower ID indicates tighter, lower-dimensional structure.

As shown in Figure~\ref{fig:intrinsicvscossim} and Table \ref{tab:id-metrics}, the 3- and 5-model groups display a steady decline in Id over rounds, consistent with semantic compression through discussion. The sharpest drop occurs between the initial round (R0) and the first exchange (R1), with smaller reductions thereafter. In contrast, the 2-model setup remains relatively stable, suggesting limited restructuring of the representational space.

\subsection{Per-Model Semantic Complexity}

Figure \ref{fig:per-model} shows per-model Id trajectories. In the 2-model setup, both LLaMA3.3 and Maverick maintain high, stable Id values. In the 3-model setup, Maverick spikes around R3 before returning to baseline, while the others remain flatter. The 5-model setup exhibits the greatest divergence: Gemma rises sharply before dropping, Deepseek fluctuates moderately, and Mistral shows two steep drops at R2 and R5, indicating that larger groups may induce greater instability and diversity in semantic complexity.

\paragraph{Pooled vs. per-model Id.}
The \textit{pooled} Id curves in Figure \ref{fig:intrinsicvscossim} are computed by concatenating all codes from all models at a given round into a single set before estimating intrinsic dimensionality. This treats the group as a unified annotator and reflects the overall diversity of the shared representational space. In contrast, the per-model curves in Figure \ref{fig:per-model} (see Appendix D) estimate Id separately for each model's codes across comments, capturing how much semantic variability each agent maintains over time.  

Averaging per-model Id values would not be equivalent to the pooled estimate: pooled Id incorporates cross-model differences \emph{within} a round, while per-model Id measures only \emph{within-model} variation. Thus, pooled Id can be lower than individual Ids when models converge on similar codes, even if each model’s own space remains internally diverse.

\subsection{Cosine Similarity vs. Intrinsic Dimension}

\begin{figure}
    \centering
    \includegraphics[width=\linewidth]{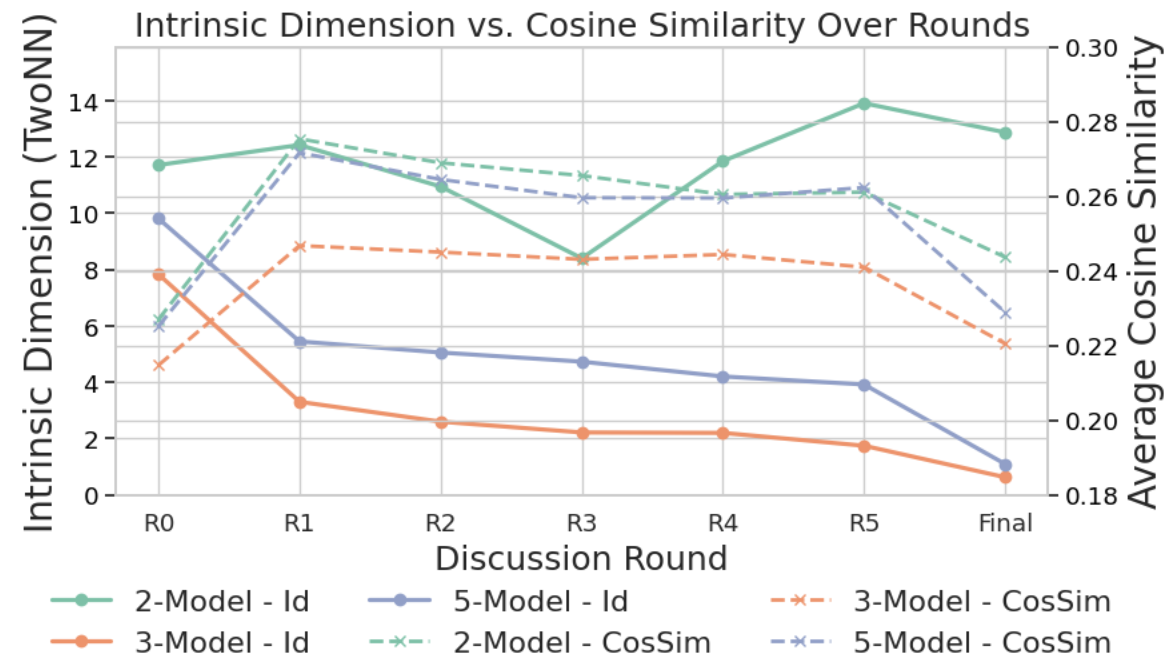}
    \caption{Comparison of intrinsic dimension (solid) and average cosine similarity (dashed) across rounds for 2-, 3-, and 5-model setups. While cosine similarity shows slight increases (surface-level alignment), intrinsic dimension reveals strong semantic compression, especially in the 5-model group. This suggests deeper convergence beyond lexical overlap.}
    \label{fig:intrinsicvscossim}
\end{figure}

To compare semantic compression with surface-level convergence, we also track average pairwise cosine similarity. While cosine similarity rises steadily in all setups and then again decreases at the final round, giving a marginally small increase in similarity overall, Id decreases much sharply (Figure~\ref{fig:intrinsicvscossim}). This divergence suggests that models become lexically closer while reducing the complexity of their outputs, supporting the idea that cosine similarity captures \textit{alignment in wording}, whereas Id reflects deeper \textit{compression of semantic space}.

\section{Structural and Affective Linguistic Features}

To capture how LLM interaction dynamics shape the linguistic, stylistic, and emotional properties of generated codes, we computed 190 discussion-level features using the ELFEN toolkit. These features span multiple linguistic dimensions including syntactic complexity, lexical diversity, readability, part-of-speech distributions, emotional valence and arousal, sensorimotor concreteness, and psycholinguistic norms.

\subsection{Affective Divergence and Emotional Breakdown}

Prompt~3, an academic-style instruction asking models to adopt a \textit{social scientist} perspective, produces the lowest ROUGE convergence of all prompt types (Table~\ref{tab:rouge_l_summary}). This suggests that interpretive ambiguity and abstraction hinder lexical alignment. The affective trajectories (Figure~\ref{fig:worstemot}, Appendix C) mirror this pattern: trust, joy, and valence drop sharply mid-discussion, while fear, sadness, and arousal rise, particularly for Maverick, which undergoes an affective collapse.  

In contrast, the highest-performing prompt (Figure~\ref{fig:bestemot}, Appendix C) shows steady affective features: trust, dominance, and positive sentiment gradually increase, and negative sentiment remains low and stable. These findings suggest that maintaining a consistent emotional tone, rather than fluctuating between affective extremes, supports shared understanding and iterative refinement.

\subsection{Structural and Lexical Differences}

We report (i) \textbf{Yule’s K} (lexical repetitiveness; higher means more repetition), (ii) \textbf{Hapax rate} (ratio of singletons; higher means more novelty), and (iii) \textbf{HDD} (Hypergeometric Distribution Diversity; robust type–token measure). We also track readability (Flesch) and syntactic depth (constituent tree depth).

Lexical and structural analyses (Figures~\ref{fig:worststruct}, \ref{fig:beststruct}, Appendix C) highlight substantial differences in discourse coherence between low- and high-performing prompts. Under low-performing conditions, LLMs display volatile behavior: Yule’s~K, hapax legomena rates, and HDD fluctuate unpredictably, while readability scores (e.g., Flesch) reach implausible extremes for some models. Deepseek and Maverick show high lexical volatility, while Gemma spikes in Yule’s~K—suggesting degeneracy or repetitive overfitting.  

Conversely, high-performing prompts yield flatter or gradually improving trends in lexical diversity and sentence complexity. Maverick and LLaMA~3.3 demonstrate increasing syntactic tree depth, lexical density, and stable POS chunking. 

\begin{table*}[t]
\centering
\small
\caption{Excerpted transcripts showing negative (left) and positive (right) influence patterns. Each turn shows the speaking model and its proposed code, with adopted lexical items in \textcolor{blue}{blue}.}
\label{tab:qualexamples}
\begin{tabular}{p{0.47\linewidth} p{0.47\linewidth}}
\toprule
\textbf{Row 0: Semantic Flattening} & \textbf{Row 30: Positive Alignment} \\
\midrule
\textbf{Maverick (R0)}: Criticizing Enabling Culture &
\textbf{Maverick (R0)}: Challenging Sexist Stereotypes \\
\textbf{LLaMA3.3 (R0)}: Compassionate Condemnation &
\textbf{LLaMA3.3 (R0)}: Challenging Stereotypical Portrayals \\
\textbf{Deepseek (R0)}: Critique of Ineffective Urban Compassion &
\textbf{Deepseek (R0)}: Challenging Sexist Stereotypes \\
\textbf{Gemma (R0)}: \textcolor{blue}{Compassion Fatigue} \& \textcolor{blue}{Blaming Policy} &
\textbf{Gemma (R0)}: Gendered Representation in Mining Media \\
\textbf{Mistral (R0)}: Misguided Compassion &
\textbf{Mistral (R0)}: Challenging Media Stereotypes: Independent Women vs. Sexualized Roles \\
\textbf{All models (Final)}: \textcolor{blue}{Exasperated Urban Compassion Fatigue \& Policy Critique} &
\textbf{All models (Final)}: \textcolor{blue}{Challenging Sexist Stereotypes in Media} \\
\bottomrule
\end{tabular}
\end{table*}

\subsection{Sensorimotor Grounding}

Psycholinguistic features from the Lancaster Norms \cite{lynott2020lancaster}  reveal nuanced differences in perceptual grounding across prompt conditions (see Figures~\ref{fig:worstpsych}, \ref{fig:bestpsych}, Appendix C). Surprisingly, the worst-performing prompt shows stable or even increasing visual sensorimotor activation in several models (e.g., Mistral, Maverick), despite concurrent declines in socialness, concreteness, and emotional coherence. This may indicate that, when alignment fails, models compensate by anchoring discourse in concrete or perceptually vivid language—a possible fallback strategy when abstract coordination breaks down.  

\begin{table*}[t]
\centering
\small
\begin{tabular}{lccccc}
\toprule
\textbf{Setup} & \textbf{Initial Id} & \textbf{Final Id} & \textbf{$\Delta$Id (Final - R0)} & \textbf{Steepest Drop} & \textbf{Drop Round} \\
\midrule
2-Model & 13.55 & 13.11 & -0.44 & -1.72 & Final \\
3-Model & 7.94  & 0.64  & -7.30 & -4.63 & R1 \\
5-Model & 7.66  & 0.42  & -7.24 & -3.75 & R1 \\
\bottomrule
\end{tabular}
\caption{Intrinsic dimension (TwoNN) metrics across different model group sizes. All setups start at round R0 and proceed through multi-round discussions. Semantic compression is most pronounced in the 3- and 5-model configurations.}
\label{tab:id-metrics}
\end{table*}

In contrast, the best-performing prompt exhibits flatter or slightly declining visual grounding over rounds, accompanied by stronger emotional calibration and lexical stability. This suggests that high-quality convergence does not require perceptual vividness; rather, it benefits from emotional and structural consistency. Overall, sensorimotor features provide useful but non-linear signals.

\section{Qualitative and Error Analysis}
\label{sec:erroranalysis}

While overall trends point to increasing convergence across rounds, individual discussion threads reveal that this process is not uniformly successful. In some cases, interaction leads to clearer, more aligned codes; in others, it produces drift, vagueness, or unnecessary complexity. To illustrate, we present two examples from the 5-model, 5-round setting: one where peer interaction improves quality, and one where it undermines it.

\paragraph{Semantic drift and oversimplification.}
In some discussions, models converge confidently on codes that are less nuanced than their starting points. For example, in one case, the initial codes were thematically diverse -\textit{``Criticizing Enabling Culture''}, \textit{``Compassionate Condemnation''}, \textit{``Misguided Compassion''}, and a detailed multi-component code from Gemma explicitly identifying both \textcolor{blue}{compassion fatigue} and \textcolor{blue}{blaming policy}. Over the discussion, these variants collapsed into a highly uniform label, \textit{``Exasperated Urban Compassion Fatigue \& Policy Critique''}, adopted by all models. While lexical similarity and ROUGE-L scores increased sharply ($+0.61$), intrinsic dimensionality dropped from $6.10$ to $1.07$, signalling reduced representational diversity. This \emph{semantic flattening} risks erasing meaningful sub-themes in pursuit of consensus.

\paragraph{Positive lexical alignment.}
Other interactions show peer influence improving conceptual clarity. In another case, the group began with semantically related but lexically varied codes such as \textit{``Challenging sexist stereotypes''}, \textit{``Gendered Representation in Mining Media''}, and \textit{``Challenging Media Stereotypes: Independent Women vs. Sexualized Roles''}. Through successive rounds, the phrase \textcolor{blue}{``Challenging Sexist Stereotypes in Media''} emerged and was adopted by all agents by the final round. Here, convergence aligned both lexical form and semantic content, increasing ROUGE-L by $+0.45$ and cosine similarity by $+0.24$, while preserving the central meaning of the original codes.

\paragraph{Illustrative excerpts.}
Table~\ref{tab:qualexamples} presents condensed transcripts from these two cases. Highlighted terms (\textcolor{blue}{blue}) indicate lexical items or phrasings introduced by one model and later adopted by others.

Overall, these cases illustrate that convergence can either reinforce accurate, semantically coherent codes or collapse diversity into overgeneralised formulations. Monitoring for semantic drift and incorporating occasional human oversight could help maintain thematic richness while benefiting from the efficiency of multi-agent LLM discussions.

\section{Discussion}

Our findings demonstrate that large language models can engage in structured, multi-turn coordination without explicit role conditioning or constraints. Across various types of prompts and group sizes, interaction yields measurable gains in lexical and semantic convergence, greater epistemic confidence, and reduced toxicity. These patterns are supported by complementary metrics, including ROUGE, self-consistency, and cross-model influence scores.

The coordination dynamics is not uniform. Influence matrices reveal stable \textit{anchor} models (e.g., Gemma, LLaMA3.3) that consistently shape peer outputs, alongside models (e.g., Deepseek) that absorb external framing. This asymmetry parallels established findings in human negotiation and persuasion, where clarity and perceived confidence can drive consensus formation \cite{moussaid2013social}.

Convergence is not solely lexical. High-performing groups maintain stable sentiment trajectories and controlled syntactic complexity, suggesting that emotional coherence and discourse stability lead to alignment. In contrast, breakdowns are marked by volatility in sentiment, lexical degeneration, and increased perceptual concreteness, which may reflect compensatory grounding strategies when alignment fails.

Geometric analysis adds a complementary perspective: intrinsic dimensionality consistently shrinks over discussion rounds, indicating semantic compression beyond surface overlap. This suggests that iterative interaction narrows the representational space toward shared conceptual frames, an effect not captured by cosine similarity or ROUGE alone. Together, these results position multi-agent LLM interaction as a process of both surface alignment and deeper conceptual convergence.

\section{Conclusion}

We presented a simulation framework for multi-agent LLM discussions in inductive qualitative coding, showing that iterative interaction leads to lexical and semantic convergence, stabilizes affect, and compresses the concept space of model outputs. These dynamics: anchoring, asymmetric influence, and progressive semantic narrowing, suggest that LLM groups can exhibit forms of collective reasoning that go beyond simple lexical alignment.

Our findings highlight the potential of structured multi-turn interaction for tasks requiring consensus and interpretability, from collaborative annotation to decision-making. Future work should extend this approach to mixed human–LLM settings, probe the stability of consensus under noisy or adversarial conditions, and analyze the effects of turn-taking, memory, and agent identity on group outcomes.

\vspace{-5pt}

\section*{Limitations}
\label{sec:limitations}

\textbf{Simulation vs.\ full agent systems.}
Agents have no explicit roles, tools, or persistent memory beyond turn summaries. This controlled, prompt-only setting isolates interaction effects but underestimates capabilities of richer multi-agent architectures.

\textbf{Sequence effects.}
We randomize the starting speaker per discussion, but order within a round remains fixed, residual sequence effects are possible.

\textbf{Round awareness.}
Prompts do not reveal how many rounds remain; nonetheless, models could infer session end from context. We do not observe final-round spikes once the order is randomized, but cannot rule out planning effects.

\textbf{Confidence proxy.}
Our lexical proxy captures \emph{expressed} assertiveness, not true epistemic certainty, and depends on lexicon coverage and normalization. Alternative normalizations (per word/sentence) and learned proxies are left for future work.

\textbf{Single dataset \& prompt sensitivity.}
We study one dataset with five prompts; patterns generalize across prompts within this dataset but may differ elsewhere. Extending to additional corpora and mixed human–LLM teams is important future work.

\textbf{External embeddings.}
Semantic metrics use external sentence embeddings; ID therefore reflects \emph{output geometry} rather than internal activations.

\section*{Ethics Statement}

The analysis in this paper focuses on the behavior of large language models in a synthetic, automated discussion setting. No human participants were involved in the generation of data or the evaluation of outputs, and no personally identifiable information is present. Moreover, we note that the models studied may reproduce or amplify stereotypes present in their pretraining data. Our toxicity analyses aim to identify such tendencies, but this work does not constitute a comprehensive bias analysis. Deploying multi-agent LLM systems for annotation in real-world settings should include human oversight, bias mitigation, and domain-specific ethical review.

\bibliography{acl_latex}

\newpage
\newpage
\newpage
\newpage\newpage
\newpage\newpage
\newpage\newpage
\newpage\newpage
\newpage\newpage
\newpage\newpage
\newpage\newpage
\newpage\newpage
\newpage\newpage
\newpage

\appendix
\newpage
\section{Simulation Procedure and Prompts}
\label{sec:appendix_algorithm}

For transparency and reproducibility, we provide in Algorithm~\ref{alg:multiturn} a detailed outline of the simulation pipeline described in Section~\ref{sec:experimental_setup}.  
The algorithm formalizes our three-phase procedure: (1) \emph{Initial code generation}, where each agent independently produces an initial qualitative code for a given input; (2) \emph{Iterative refinement}, where agents take turns updating their code in response to accumulated conversational memory built from one-sentence summaries of all prior turns; and (3) \emph{Final synthesis}, where each agent generates a concluding code based on the full shared history.  
This design ensures controlled turn-taking, scalable context management, and the absence of any external memory modules or finetuning, relying solely on prompt-based inference.

Table~\ref{tab:prompts} lists the five prompt templates used to produce discussions, rotated one after another. 
The prompts range from formal thematic analysis instructions to more concise, informal formulations, enabling us to test the effect of input framing on the group coding dynamics.

\newpage

\begin{algorithm}[t]
\caption{Multi-agent Iterative Coding Simulation}
\footnotesize
\label{alg:multiturn}
\begin{algorithmic}[1]
\Require Models $M = \{m_1, \dots, m_k\}$, number of rounds $R$, prompt $p$, dataset $\mathcal{D}$ of $n$ items
\For{each item $x \in \mathcal{D}$}
    \State \textbf{Phase 1: Initial code generation}
    \For{each agent $m_i \in M$}
        \State Generate initial code $c_i^{(0)} \gets m_i(p, x)$
        \State Summarize to one-sentence $s_i^{(0)} \gets \text{summarize}(c_i^{(0)})$
    \EndFor
    \State Initialize conversational memory $C^{(0)} \gets [s_1^{(0)}, \dots, s_k^{(0)}]$
    \State \textbf{Phase 2: Iterative refinement}
    \For{round $r = 1$ to $R$}
        \For{each agent $m_i$ in fixed turn order}
            \State Context $\gets C^{(r-1)}$ plus current item $x$
            \State Generate refinement $c_i^{(r)} \gets m_i(p, x, \text{context})$
            \State Summarize $s_i^{(r)} \gets \text{summarize}(c_i^{(r)})$
            \State Append $s_i^{(r)}$ to conversational memory $C^{(r)}$
        \EndFor
    \EndFor
    \State \textbf{Phase 3: Final synthesis}
    \For{each agent $m_i \in M$}
        \State Given $C^{(R)}$, produce final synthesis code $c_i^{\text{final}}$
    \EndFor
\EndFor
\end{algorithmic}
\end{algorithm}

\section{Confidence Score Computation}

To quantify expressions of certainty and hedging in model utterances, we define two lexicons: one for certainty cues and one for hedging cues. These lexicons are manually curated and include common adverbs, modal verbs, and multi-word phrases associated with either assertive or uncertain language.

\subsection*{Certainty Lexicon}
The certainty lexicon C includes terms such as:

\begin{quote}
\textit{definitely, must, undoubtedly, always, clearly, certainly, absolutely, without a doubt, unquestionably, conclusively, positively, with certainty, no doubt, undeniably, strongly}, etc.
\end{quote}

A full list contains 65 expressions indicative of strong epistemic commitment or assertive framing.

\subsection*{Hedging Lexicon}
The hedging lexicon H includes terms such as:

\begin{quote}
\textit{might, possibly, could, likely, seems, apparently, perhaps, maybe, presumably, arguably, supposedly, relatively, somewhat, in theory, reportedly, one might argue, from what I gather, I guess}, etc.
\end{quote}

This list includes 70 phrases commonly associated with epistemic uncertainty, mitigation, or speculative reasoning.

\subsection*{Confidence Score Formula}
Given a tokenized utterance, we compute the confidence score as:

\begin{equation}
\text{Confidence}(t) = \frac{\ \text{C terms} - \ \text{H terms}}{\text{N tokens in } t}
\end{equation}

This produces a normalized score capturing the relative assertiveness of an utterance, where higher values indicate stronger epistemic commitment and lower (or negative) values indicate hedging or uncertainty.

\begin{figure}
    \centering
    \includegraphics[width=0.8\linewidth]{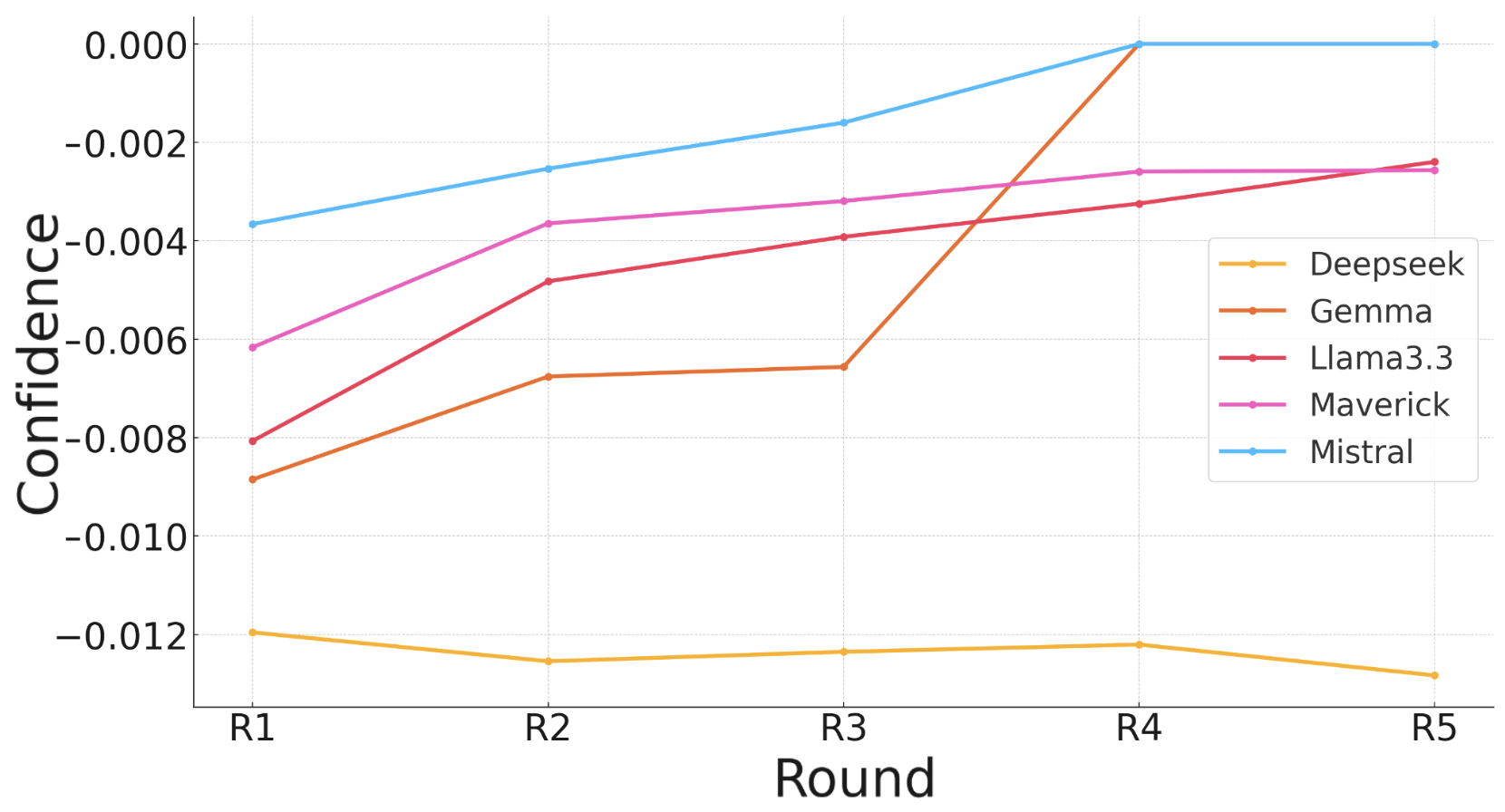}
    \caption{Confidence Score Trajectories Across Rounds.}
    \label{fig:confidence}
\end{figure}

\begin{figure}
    \centering
    \includegraphics[width=0.8\linewidth]{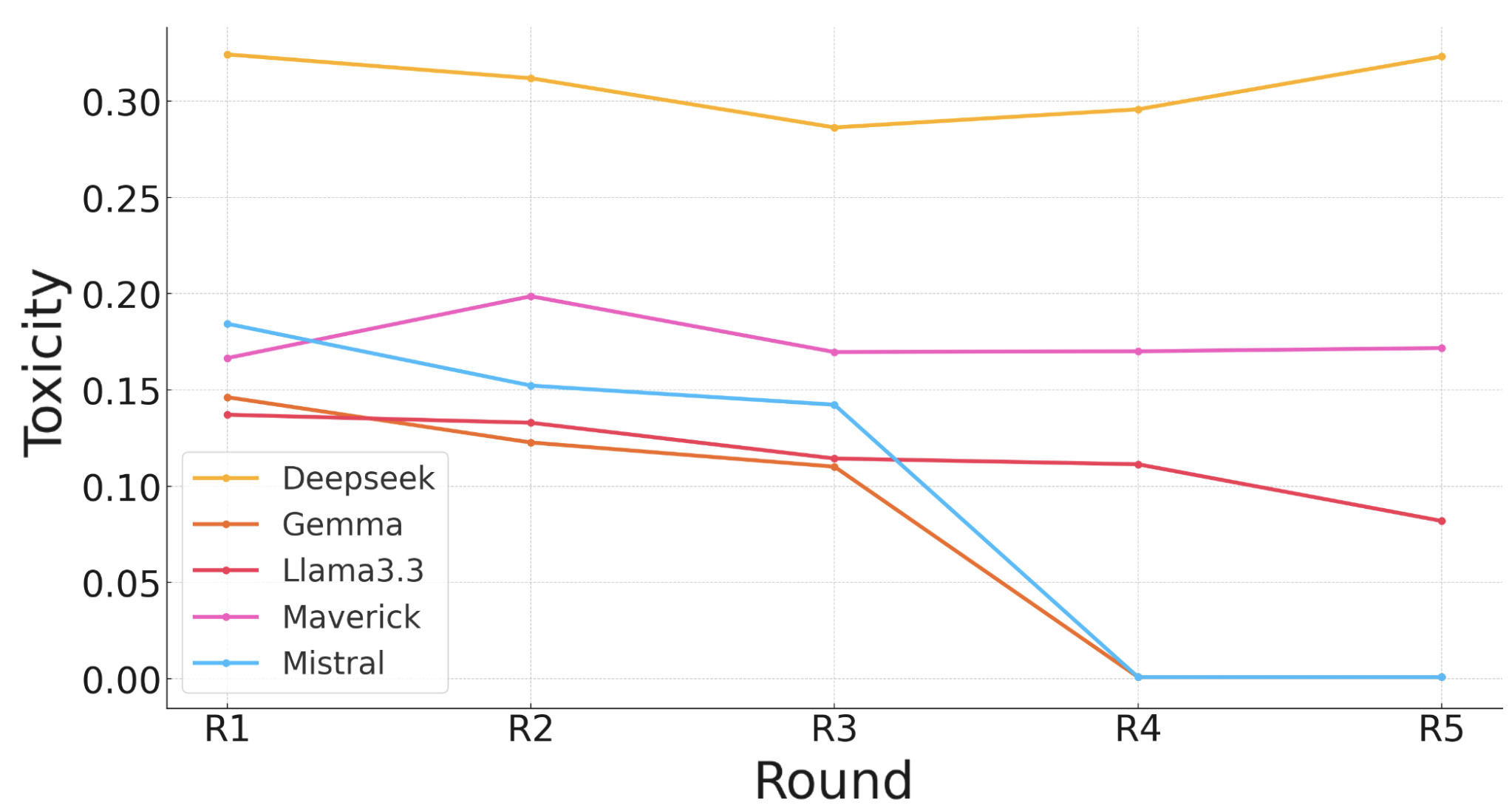}
    \caption{Toxicity Trends Across Rounds. Average toxicity scores for each model over five rounds, as measured by a zero-shot classifier finetuned for social bias and toxicity detection.}
    \label{fig:toxic}
\end{figure}

\section{Semantic Influence}

\begin{figure*}
    \centering
    \includegraphics[width=\linewidth]{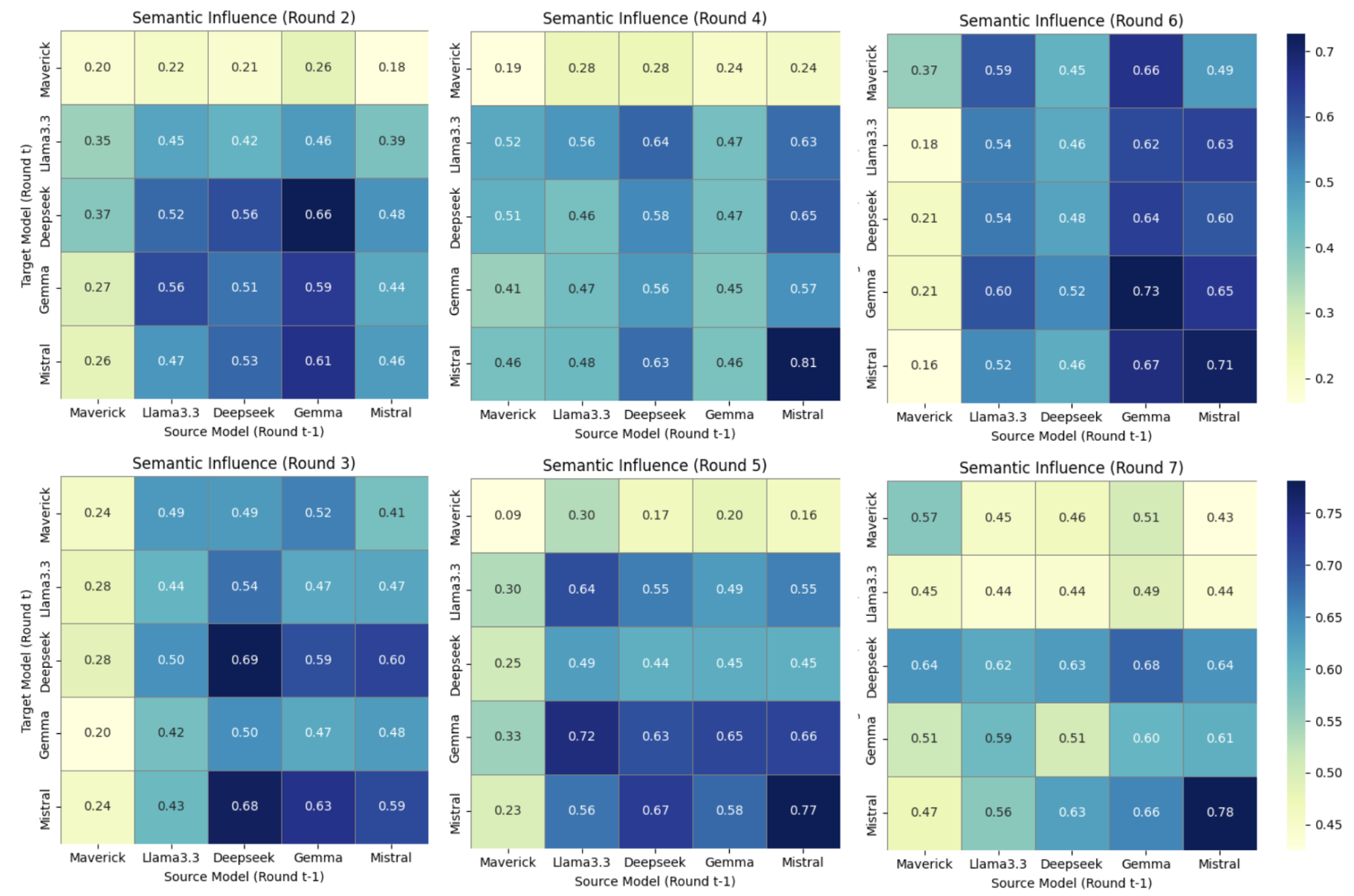}
    \caption{Semantic Influence Matrices Across Discussion Rounds.
Each heatmap represents the average pairwise cosine similarity between a target model’s code at round t and all source models’ codes from round t–1, computed using MiniLM embeddings. Diagonal values capture self-influence (semantic consistency over time), while off-diagonal values indicate cross-model influence (semantic alignment to peers). Over time, influence intensifies, especially for Mistral, Gemma, and Deepseek, reflecting increasing inter-model convergence and shared framing. Notably, Llama3.3 and Deepseek emerge as consistent semantic sources, while Maverick gradually stabilizes after early-round volatility.}
    \label{fig:semantic-influence}
\end{figure*}

\begin{figure*}
    \centering
    \includegraphics[width=\linewidth]{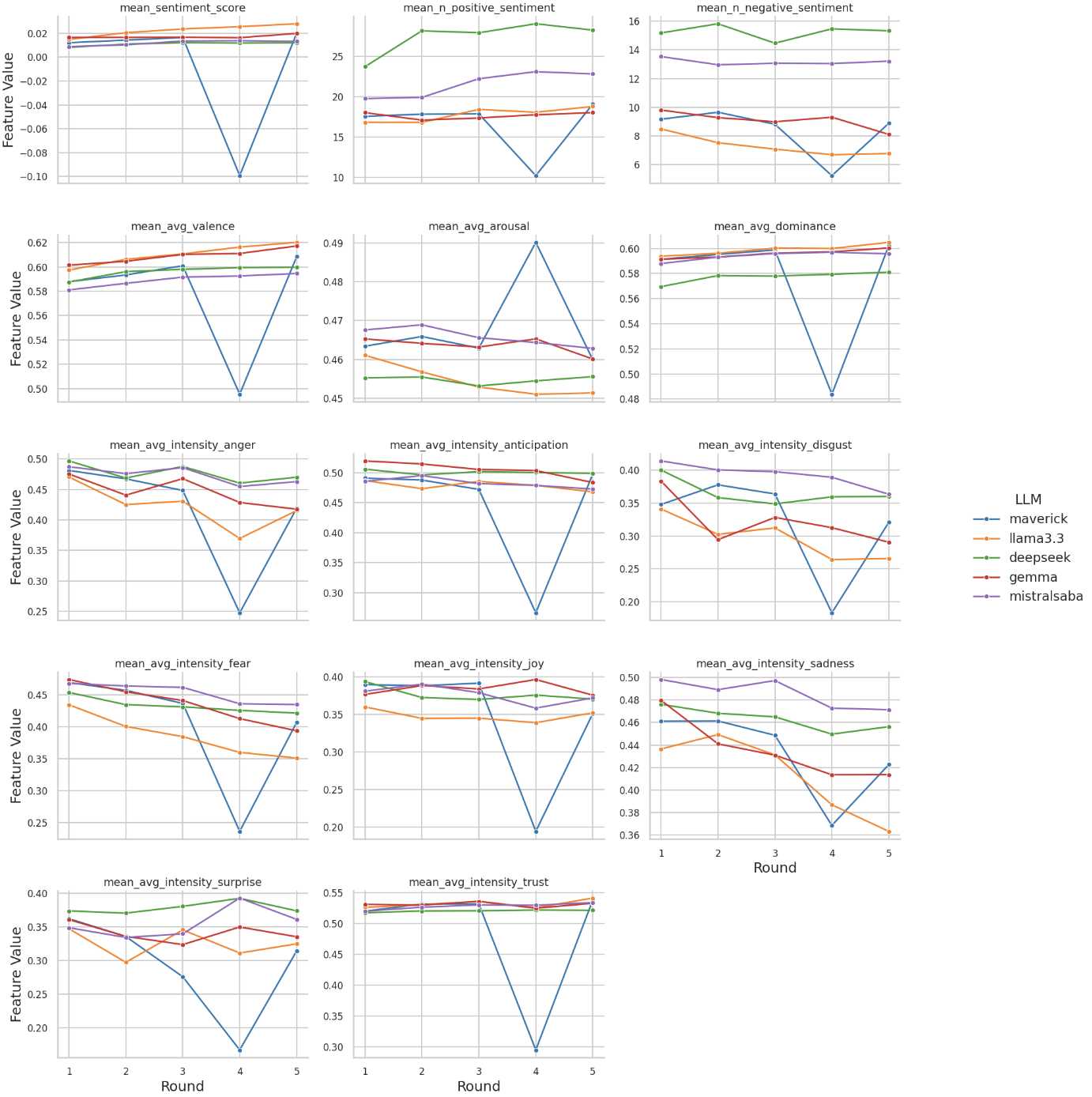}
    \caption{Affective feature trajectories for the lowest-performing prompt. Sentiment, trust, and valence drop sharply mid-discussion, while fear, sadness, and arousal increase—indicating emotional instability and breakdown in cooperative framing.}
    \label{fig:worstemot}
\end{figure*}

\begin{figure*}
    \centering
    \includegraphics[width=\linewidth]{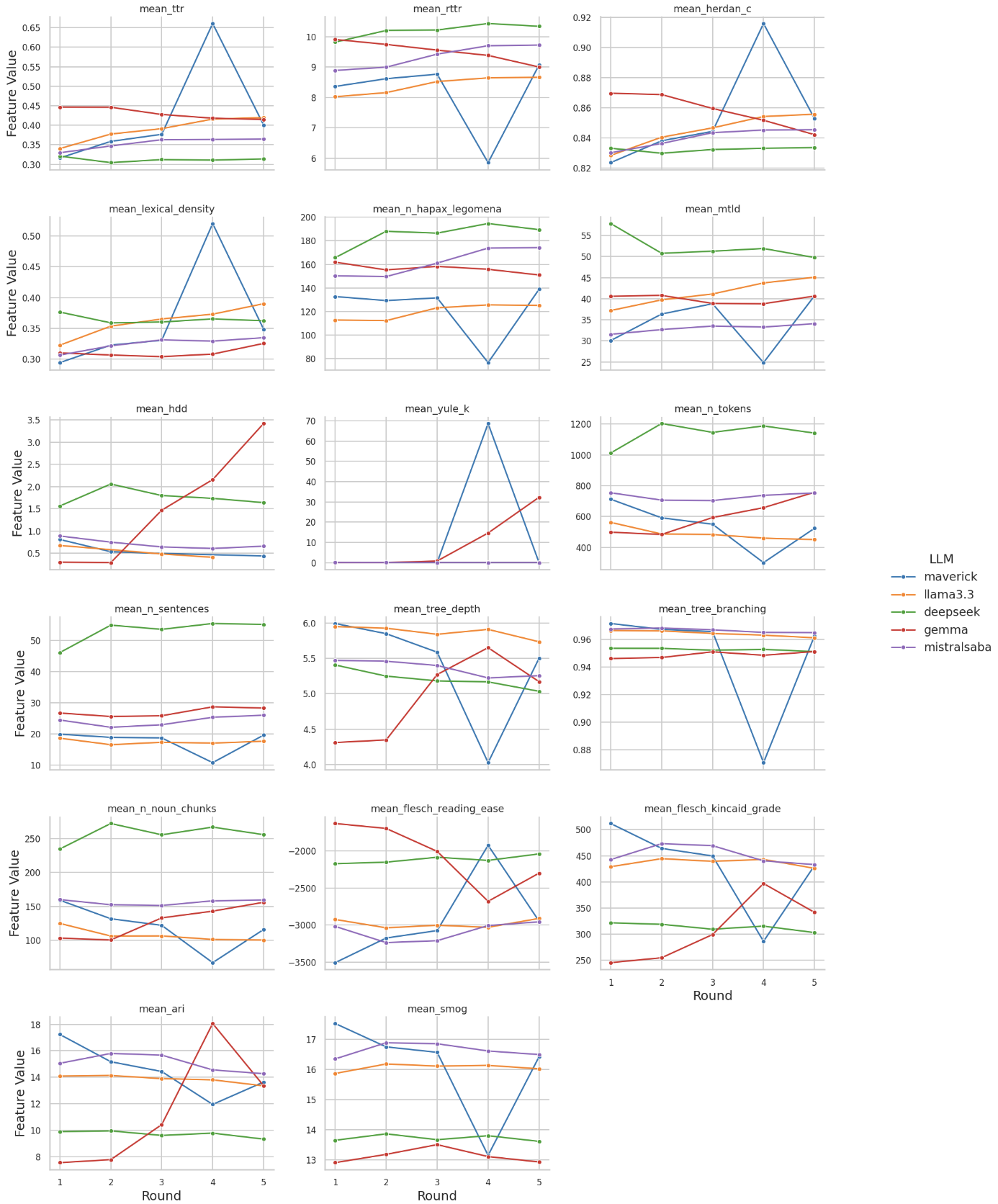}
    \caption{Lexical and structural metrics for the lowest-performing prompt. Measures such as Yule’s~K, hapax legomena, and readability fluctuate erratically, with some models exhibiting lexical degeneration or overfitting-like repetition.}
    \label{fig:worststruct}
\end{figure*}

\begin{figure*}
    \centering
    \includegraphics[width=\linewidth]{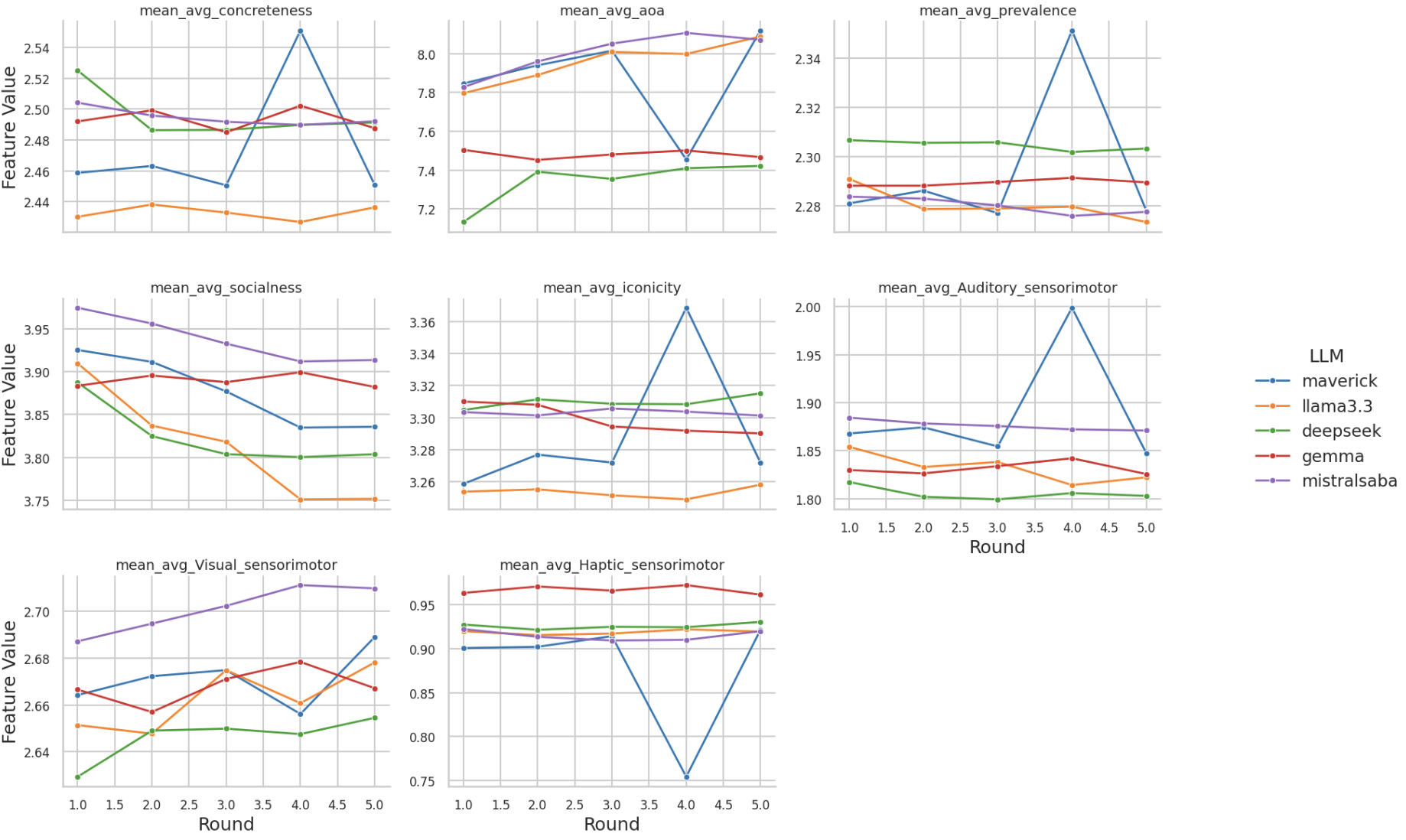}
    \caption{Psycholinguistic features for the lowest-performing prompt. Despite emotional and structural drift, some models show increased visual grounding, suggesting a compensatory shift toward concrete, perceptually vivid language when abstract alignment fails.}
    \label{fig:worstpsych}
\end{figure*}

\begin{figure*}
    \centering
    \includegraphics[width=\linewidth]{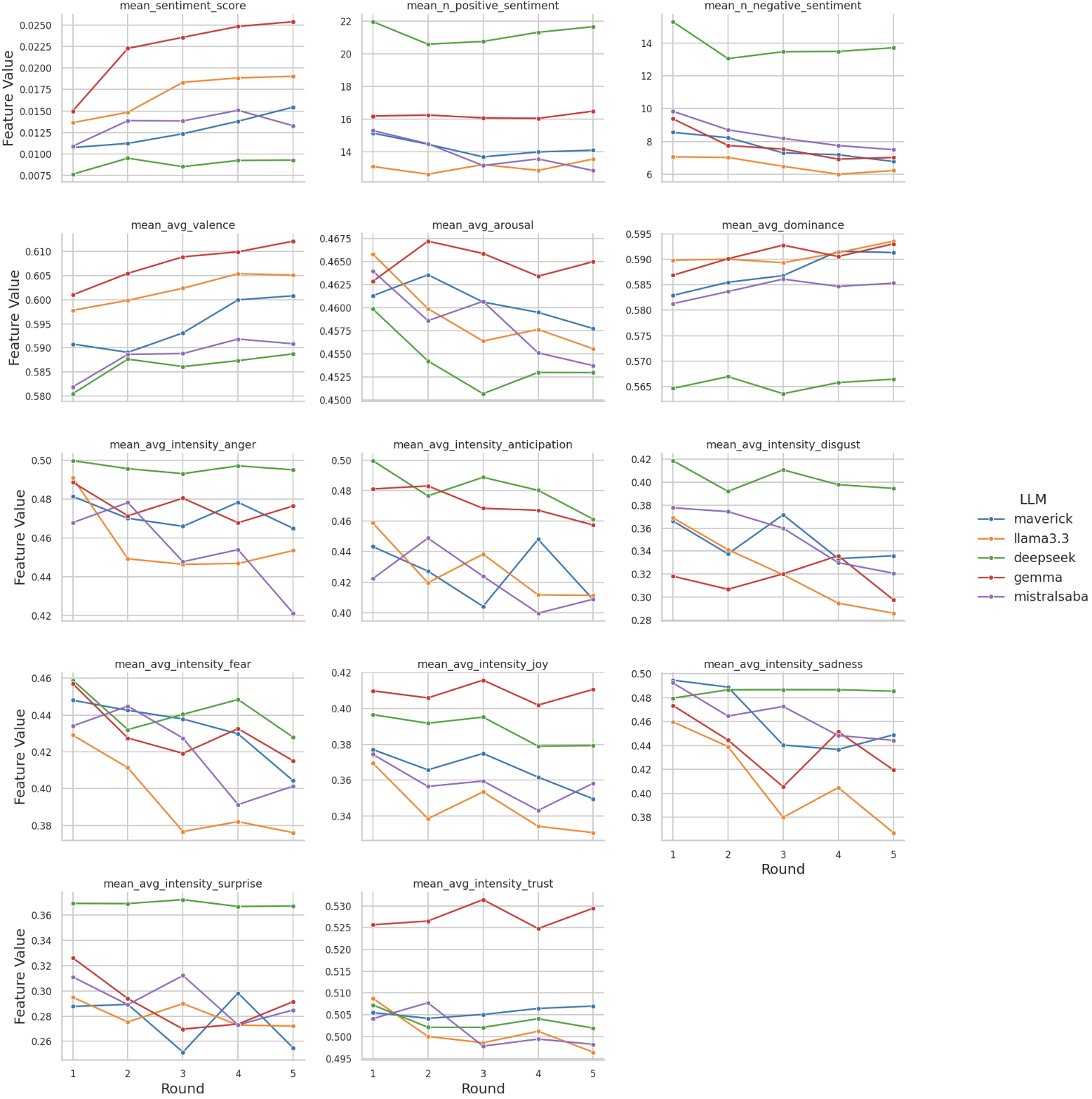}
    \caption{Affective feature trajectories for the highest-performing prompt. Trust, dominance, and positive sentiment steadily rise, while negative affect remains low—indicating stable emotional regulation and cooperative tone.}
    \label{fig:bestemot}
\end{figure*}

\begin{figure*}
    \centering
    \includegraphics[width=\linewidth]{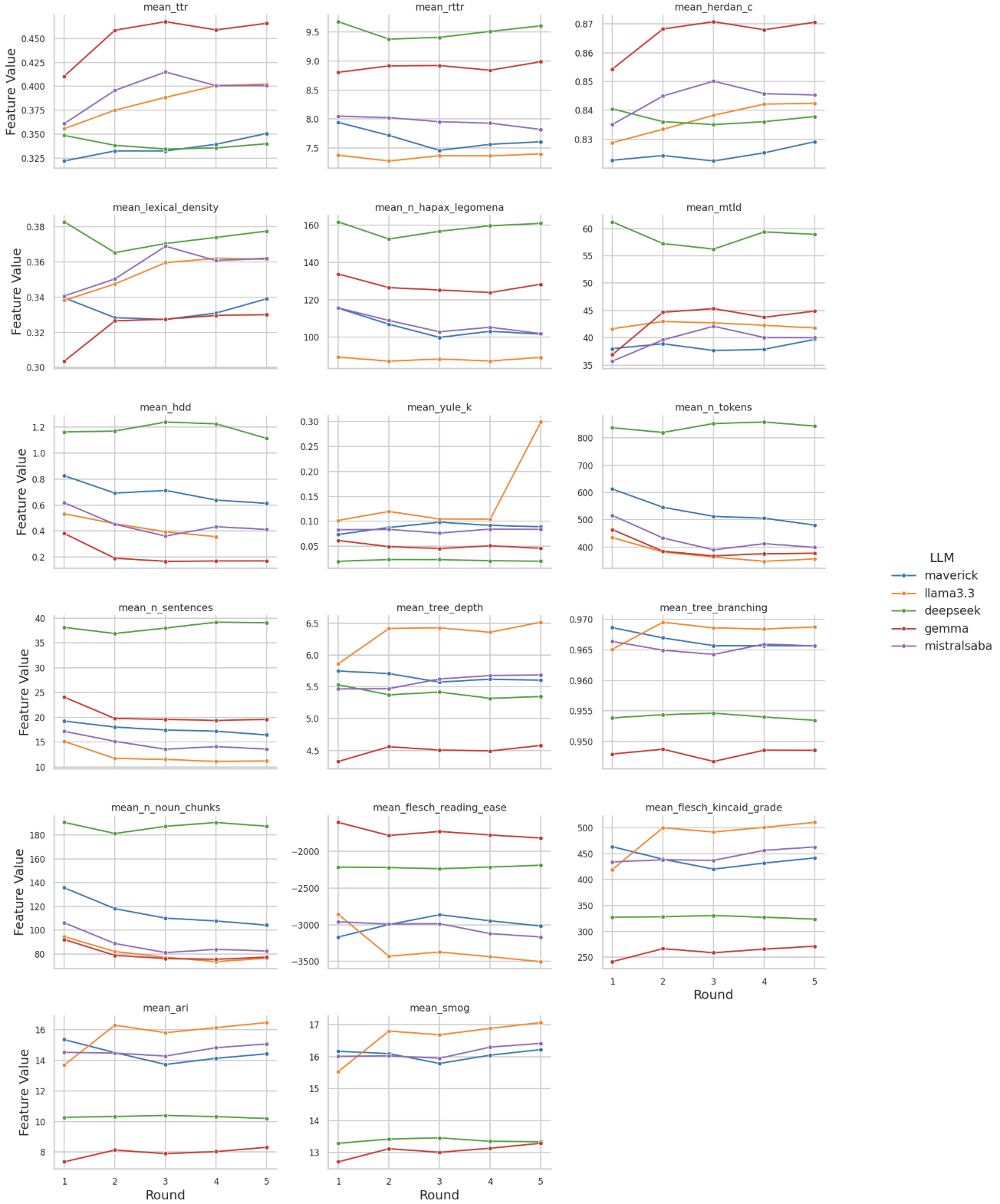}
    \caption{Lexical and structural metrics for the highest-performing prompt. Trends in lexical diversity, syntactic depth, and token economy are stable or gradually improving, reflecting sustained discourse coherence and minimal drift.}
    \label{fig:beststruct}
\end{figure*}

\begin{figure*}
    \centering
    \includegraphics[width=\linewidth]{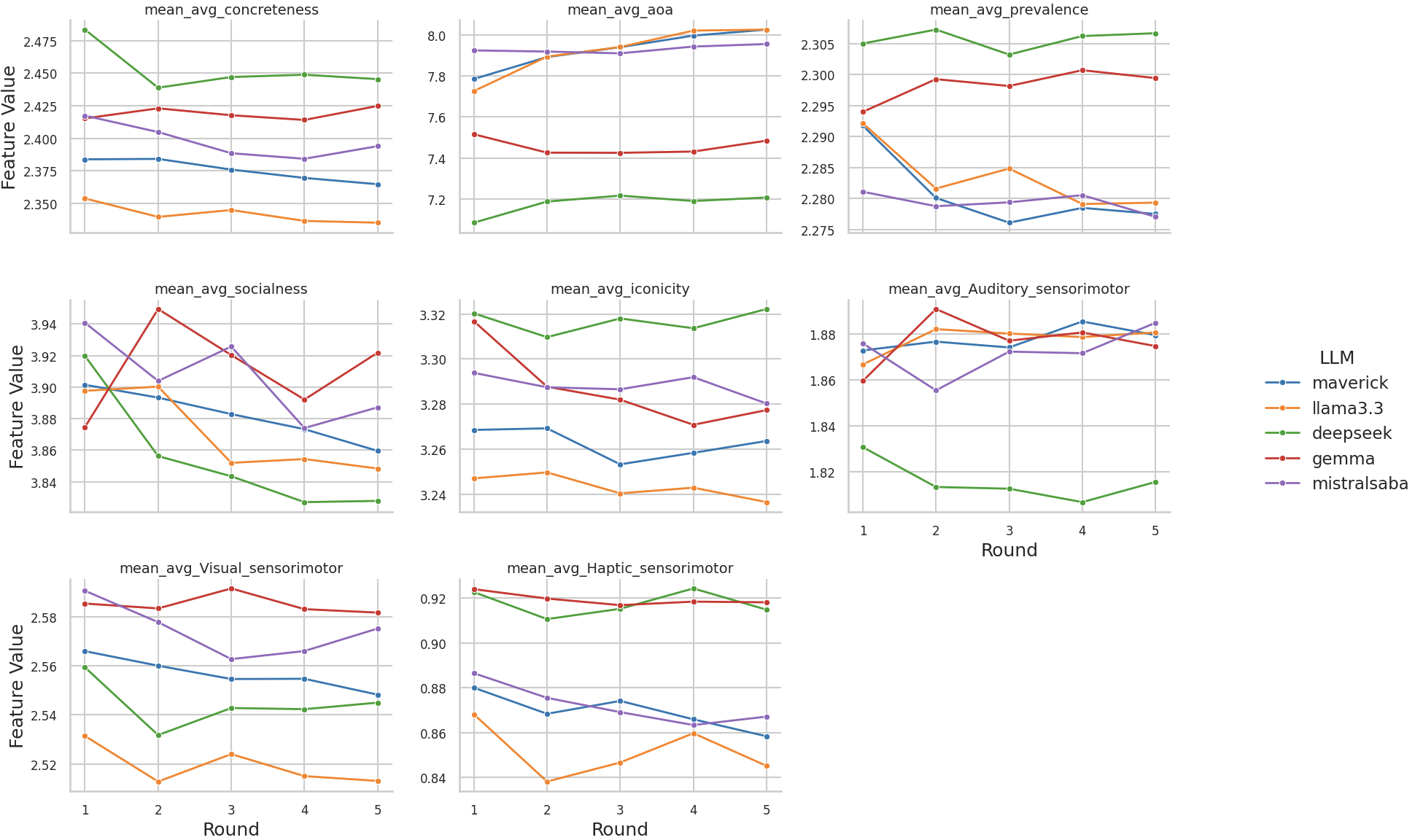}
    \caption{Psycholinguistic features for the highest-performing prompt. Visual grounding remains flat or slightly declines, while emotional and structural stability persist—suggesting effective alignment without reliance on increased concreteness.}
    \label{fig:bestpsych}
\end{figure*}

\section{Intrinsic Dimension Metrics}

\begin{figure*}
    \centering
    \includegraphics[width=0.6\linewidth]{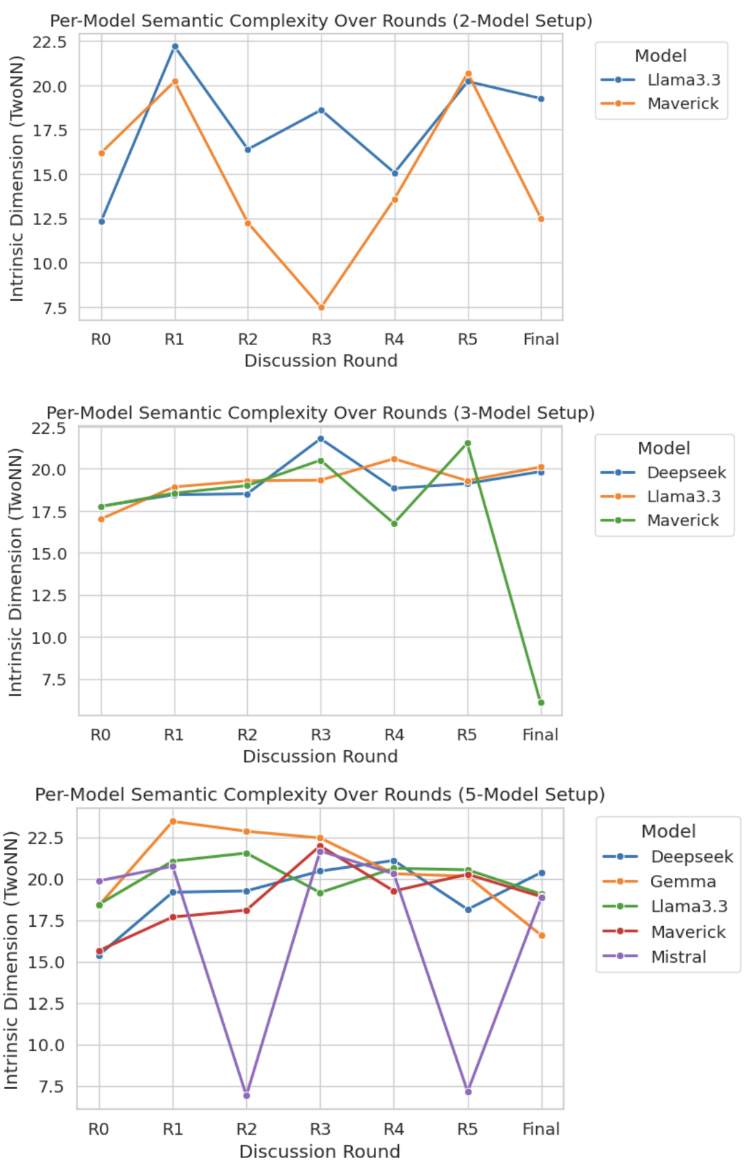}
    \caption{Per-model semantic complexity across rounds, measured via intrinsic dimension (TwoNN) of each model’s generated codes. 
}
    \label{fig:per-model}
\end{figure*}

\appendix

\end{document}